\documentclass[journal]{IEEEtran} %

\usepackage[style=ieee,dashed=false]{biblatex}
\usepackage{array}
\usepackage{graphicx}
\usepackage{amsmath,amssymb,amsthm, amsfonts,bm}
\usepackage{gensymb}
\usepackage{resizegather}
\usepackage{tabularx} %
\usepackage{nidanfloat}    %
\usepackage{etoolbox}
\newcommand{\density}{\ensuremath{{D}}} %
\newcommand{\p}{\density} %
\newcommand{\numSamples}{\ensuremath{N}} %
\newcommand{\N}{\numSamples} %

\newcommand{\occluder}{\ensuremath{O}}

\newcommand{\Var}{\operatorname{Var}}

\newcommand{\img}{\ensuremath{S}} 
\newcommand{\R}{\img}  %
\newcommand{\noise}{\occluder}

\newcommand{\X}{\ensuremath{X}}  %
\newcommand{\E}{\operatorname{E}}
\newcommand{\Ex}[1]{\ensuremath{\E[#1]}}

\newcommand{\sigmaOccl}{\ensuremath{\sigma_{o}^2}} 
\newcommand{\sigmaSign}{\ensuremath{\sigma_{s}^2}} 
\newcommand{\meanOccl}{\ensuremath{\mu_{o}}} 
\newcommand{\meanSign}{\ensuremath{\mu_{s}}} 
\newcommand{\OneZi}{(1\!-\!Z_{i})}%
\newcommand{\OneD}{(1\!-\!\p)}%

\newcommand{\appendixMath}[1]{\newline{}\scalebox{.798}{\parbox{1.25\linewidth}{#1}}\newline{}}

\begin{document}

\title{Pose Error Reduction for Focus Enhancement in Thermal Synthetic Aperture Visualization}

\author{Indrajit~Kurmi,
        David~C.~Schedl,
        and~Oliver~Bimber,~\IEEEmembership{Senior~Member,~IEEE}%
\thanks{I. Kurmi, D. Schedl, and O. Bimber are with the Computer Science Department of the Johannes Kepler University, Linz, Austria, e-mail: firstname.lastname@jku.at}%

\thanks{Manuscript received XXX XXX, 2020; revised XXX XXX, XXX.} \vspace{-1.5em}} %

\markboth{IEEE GEOSCIENCE AND REMOTE SENSING LETTERS, VOL. XX, NO. X, XXX 2020}%
{Kurmi \MakeLowercase{\textit{et al.}}: Airborne Optical Sectioning for Search and Rescue}
\maketitle
\begin{abstract}
Airborne  optical  sectioning,  an  effective  aerial  synthetic  aperture imaging technique  for  revealing  artifacts  occluded  by forests, requires precise measurements of drone poses. In this article we present a new approach for reducing pose estimation errors beyond the possibilities of conventional Perspective-n-Point solutions by considering the underlying optimization as a focusing problem. %
We present an efficient image integration technique, which also reduces the parameter search space to achieve realistic processing times, and improves the quality of resulting synthetic integral images. 
\end{abstract}
\begin{IEEEkeywords}
Image Processing and Computer Vision, Enhancement.
\end{IEEEkeywords}
\vspace{-1.0em} 
\IEEEpeerreviewmaketitle
\section{Introduction}
\IEEEPARstart{A}{irborne} optical sectioning (AOS) \cite{Kurmi2018,Kurmi2019TAOS,Kurmi2019Sensor,Bimber2019IEEECGA,Kurmi2020GeoScience, David2020} is an effective aerial synthetic aperture imaging technique for revealing artifacts which are otherwise occluded through dense forest and remain invisible in individual images. For applications such as search and rescue (SAR) \cite{Kurmi2019TAOS} and wild life observation \cite{David2020}, %
AOS acquires the fractional heat signal radiated from occluded targets as an unstructured thermal light-field recorded with a camera drone (Fig.~\ref{fig:Fig0CameraPose}).
Capturing over a wide synthetic aperture area (30–100 m diameter) support optical sectioning by image integration \cite{Kurmi2018}. \vspace{-1.5em}
\begin{figure*}[!hb]
    \vspace{-0.5em}\includegraphics[width=0.99\linewidth]{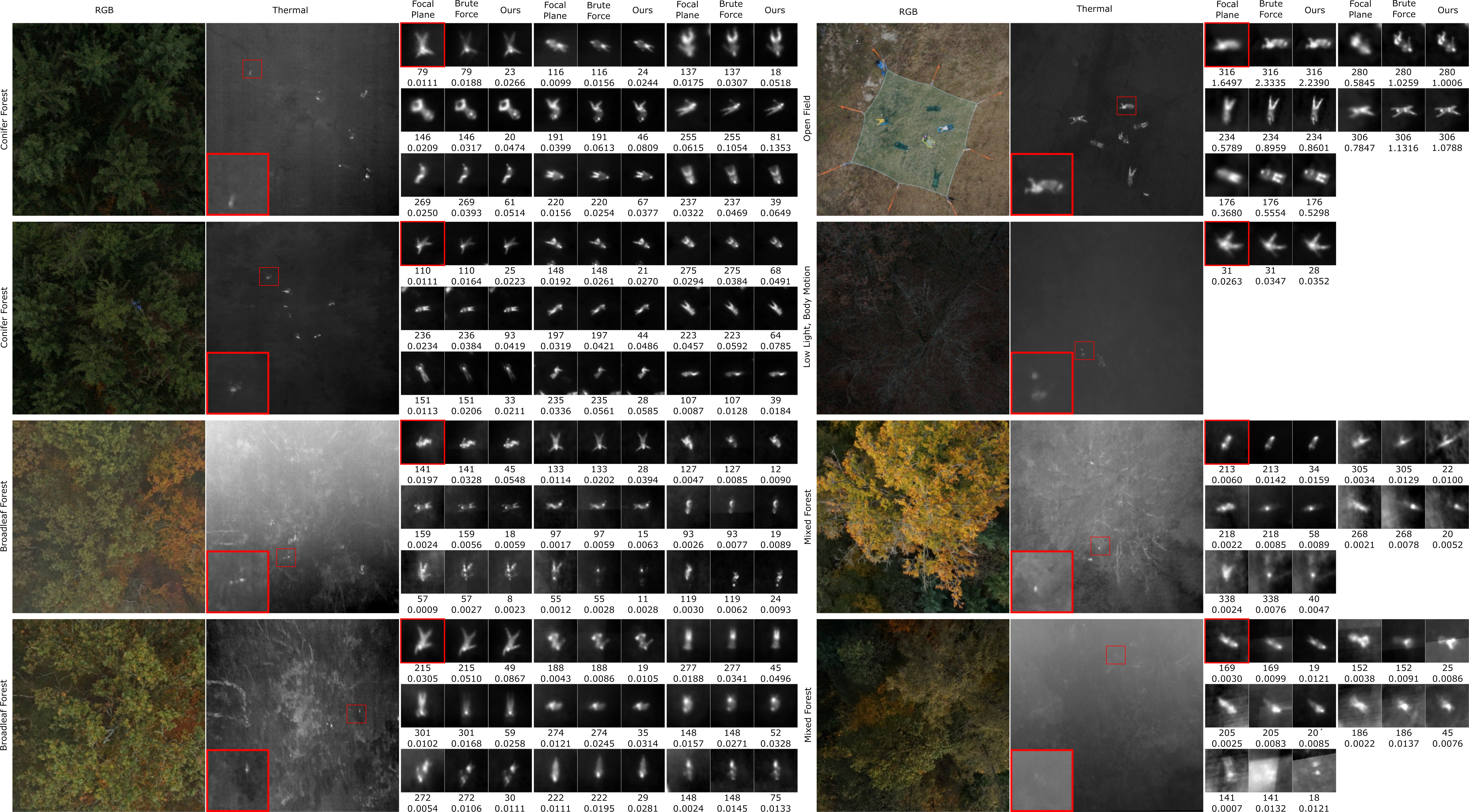}
    \vspace{-0.5em}\caption{ Results of pose error reduction for 8 different scenes and varying recording conditions (exemplary RGB and thermal images are shown). We compare {\it focal plane} optimization~\cite{Kurmi2020GeoScience}, {\it brute force} search, and {\it our} approach for 52 test cases (i.e., humans). The number of integrated images ($N$) as well as the resulting normalized variance ($N \cdot \Var[\X]$) are displayed below each test case. %
    Insets in the thermal images show positions and closeups for one exemplary human per scene (highlighted in red). %
    }
    \label{fig:Fig1Completeresult}
    \vspace{-1.8em}
\end{figure*}
The  integral  images  (representing  the  signal  of  a  wide-aperture  sensor  measurement)  are  computed  by  registering and averaging all single image recordings with respect to the drone's three-dimensional poses and a defined synthetic focal plane. %
RGB images in visible waveband (Sony Alpha 6000 RGB camera) are acquired for drone pose estimation and thermal images (Flir Vue Pro; 7.5-13.5 $\mu m$ spectral band) for integration. Single image recordings (both thermal and RGB) are stored on the camera's internal memory cards and are downloaded for further processing after flights. 
After integration, targets at the focal plane will appear clear and sharp while occluders at greater distances will be strongly blurred and vanish. Thus, shifting focus by adjusting the focal plane interactively or automatically \cite{Kurmi2020GeoScience} towards its optimal position and orientation near the ground allows optical slicing through forest and the discovery and inspection of concealed targets, such as human bodies or animals. 

AOS requires precise measurements of the drone's pose for each captured thermal image. 
Currently it is implemented by applying state-of-the-art computer vision algorithms using general-purpose structure-from-motion and multi-view stereo pipeline \cite{schonberger2016sfm,schoenberger2016mvs} to the high resolution RGB images. However, a recent study \cite{Chin2018} has shown that primarily estimating or fitting a model to data with outliers is an NP-hard problem. Thus, pose estimation remains error prone and resulting mis-registrations of single recordings leads to defocus in AOS integral images.
Note that, due to their low resolution, high noise level, low contrast, and poor texture of natural scenes, thermal images do not provide usable image features for reliable pose estimation.
 
Various approaches for correcting the remaining pose-errors have been proposed, such as optimizations with either imposing structural constraints \cite{Li2018,Ovechkin2018,Hsiao2018}, by utilizing localized depth maps \cite{Shin2018,Shamwell2019}, or by fusing different sensors \cite{Mur-Artal2017,Qin2018}.

In this article, we present a different approach for reducing remaining pose errors by considering the underlying optimization as a focusing problem instead of solving the Perspective-n-Point (PnP) problem \cite{Chin2018}.
PnP solutions \cite{Li2018,Ovechkin2018,Hsiao2018,Shin2018,Shamwell2019,Mur-Artal2017,Qin2018} require robust feature detection which is challenging in low-resolution thermal images and infeasible in the presence of strong occlusions. %
Thus, we solve the focusing problem by optimizing 3 or 6 pose parameters per image using simple gray level variance (GLV)\cite{Firestone1991} (which does not rely on any image features) as an error metric. %
We have already proven in \cite{Kurmi2020GeoScience}, that GLV is invariant to occlusion and is therefore (in contrast to traditional \mbox{gradient-,} Laplacian-, and wavelet-based metrics reviewed in \cite{Pertuz2013}) an ideal visibility metric for AOS thermal integral images.

Figure \ref{fig:Fig1Completeresult} illustrates AOS results when the synthetic focal plane is globally optimized as presented in \cite{Kurmi2020GeoScience} (focal plane), and the improved results when our new pose error reduction approach is applied to either 6D (brute force) or 3D (ours) pose parameters. For the latter, resulting integral images reveal more details as they are better focused and have a higher contrast.   

In the following sections, we will discuss effects of errors in different pose parameters and various integration strategies leading to our final optimization approach with reduced parameter search space and early stopping. We will demonstrate that our approach improves the quality of integral images (compared to state-of-the-art pose estimation methods) while also being significantly faster and more efficient than the brute force optimization of all pose parameters.
\vspace{-0.8em}
\section{The Effect of Pose Errors}\label{section:EffectofPoseErrors}
By applying a conventional geometric camera model, a scene point $P = (X,Y,Z,1)^{T} $ in world coordinates is projected to the image point $p = (x,y)^{T}$ in coordinates of the camera's image plane by the projection matrix $M$ ($p=1/zMP$), where $M = K (R \, t)$.
\begin{figure}[!htbp]
    \centering
    \includegraphics[width=0.6\linewidth]{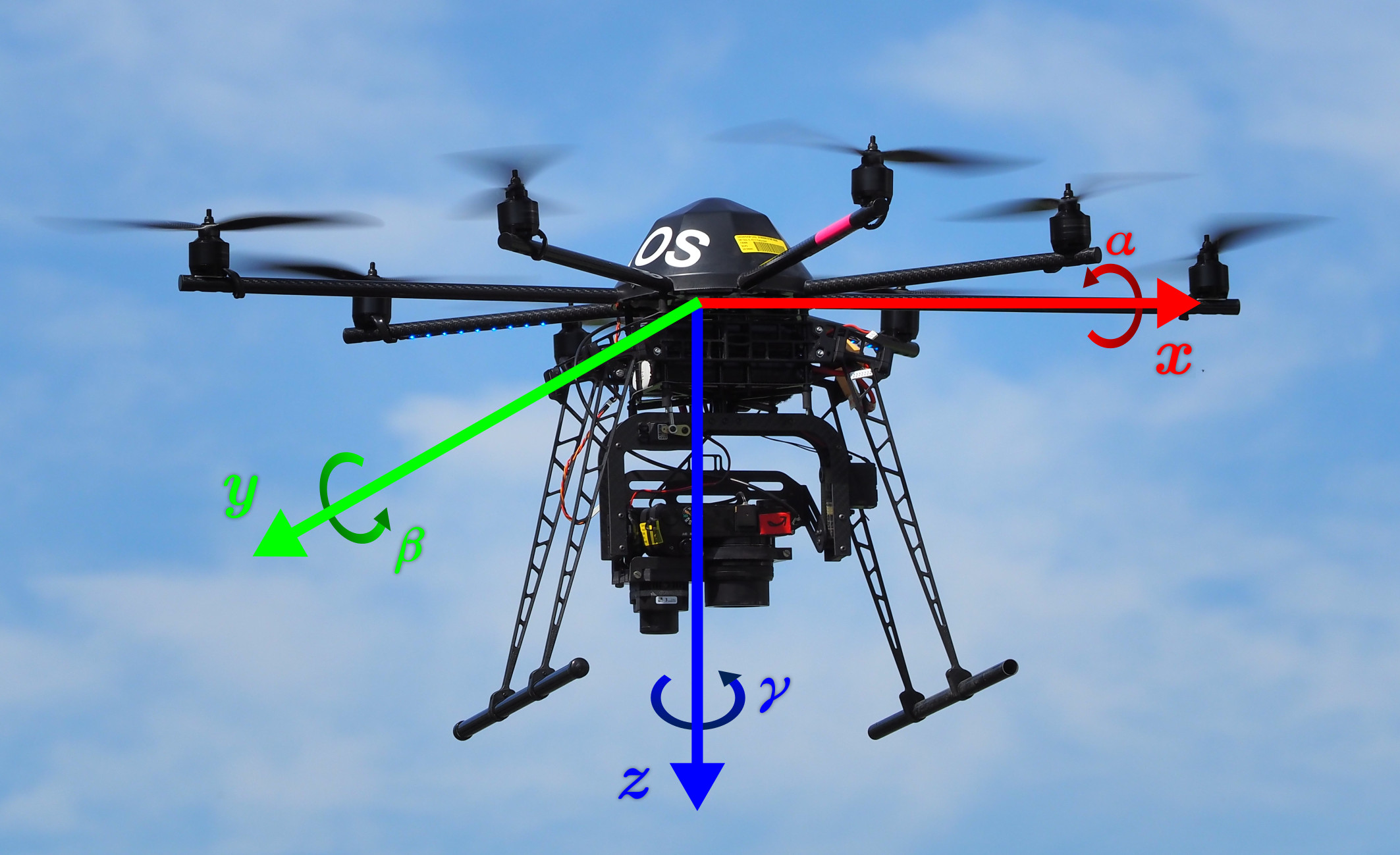}
    \vspace{-0.6em}\caption{Our MikroKopter OktoXL 6S12 octocopter equipped with a Flir Vue Pro thermal camera
(9 mm fixed focal length lens, 14 bit tonal range covering
a 7.5-13.5 $\mu m$ spectral band) and a Sony Alpha 6000 RGB
camera (16-50 $mm$ lens, set to infinite focus) and its local coordinate system. The transformation to the global coordinate system is given by the extrinsic transformation parameters ($t_x, t_y,t_z,\alpha, \beta, \gamma$).}
    \label{fig:Fig0CameraPose}
    \vspace{-1.8em}
\end{figure}
While the camera's intrinsic matrix $K$ is constant for calibrated fixed focal length cameras, the extrinsic matrix ($R \, t$) encodes the pose in form of a position column vector $t$ and an rotation matrix $R$ (see Fig.~\ref{fig:Fig0CameraPose}). %
Note that $(R \, t)$ denotes the column-wise concatenation of a matrix and vector. %
The coefficients of $R$ are composed of the three rotation parameters $\alpha$ (rotation around x-axis / pitch axis), $\beta$ (rotation around y-axis / roll axis), $\gamma$ (rotation around z-axis / yaw axis). %
To calculate $R$ we use $R = R_x(\alpha) R_y(\beta) R_z(\gamma)$, where $R_x, R_y, R_z$ are individual rotation matrices about axes x,y,z ($z$-$y$-$x$ extrinsic active rotation). %
The coefficients of $t$ are the three translation parameters ($t_x, t_y,t_z$) along pitch, roll, and yaw axes.

\begin{figure}[!bh]
    \centering
    \includegraphics[width=0.95\linewidth]{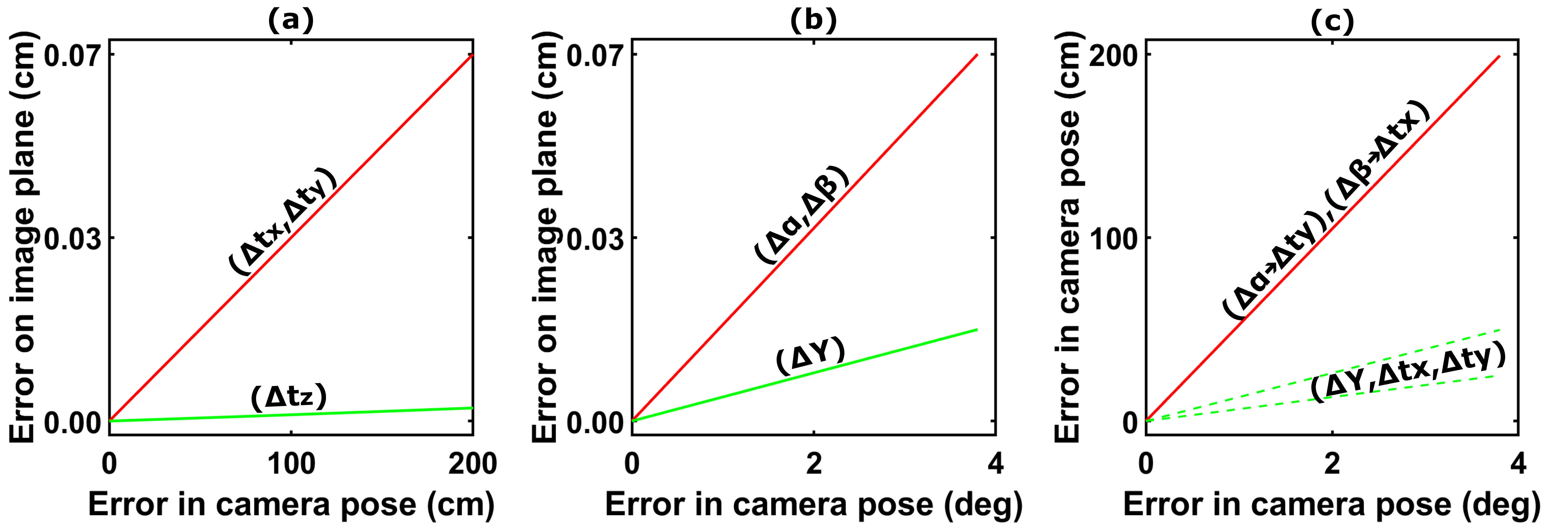}
    \vspace{-0.4em}\caption{%
    Effect of pose estimation error on image plane error. In this simulation, we assume that $t_{z}\gg Z$ and that $t_{z}=30$m. (a) Image plane error for $\Delta t_{z}$ (green) and for $\Delta t_{x},\Delta t_{y}$ (red). (b) Image plane error for $\Delta \gamma$ (green) and for $\Delta \alpha,\Delta \beta$ (red). (c) Same image plane error caused by $\Delta t_{x}$,$\Delta t_{y}$ or by corresponding $\Delta \alpha,\Delta \beta$ (red). Image plane error of two different points caused by $\Delta t_{x}$,$\Delta t_{y}$ show that no constant corresponding $\Delta \gamma$ exists for all points (dashed green).%
    }
    \label{fig:Fig2EffectPoseError}
    \vspace{-0.8em}
\end{figure}

Analyzing how much (on average) pose estimation error effects projection error on the image plane reveals that not all extrinsic parameters have the same impact. 
Figure \ref{fig:Fig2EffectPoseError} illustrate these dependencies under the assumption that the camera is located sufficiently far away from the recorded object ($t_{z}\gg Z$) to ignore perspective distortion.
While a pose error ($\Delta t_{z}$) in $t_z$ leads to no significant error on the image plane, pose errors ($\Delta t_{x},\Delta t_{y}$) in $t_{x}$ and $t_{y}$ result in notable projection errors (Fig. \ref{fig:Fig2EffectPoseError}a). 
Rotation errors along pitch and roll axes ($\Delta \alpha$ and $\Delta \beta$) cause more projection error than a rotation error along the yaw axis $\Delta \gamma$ (Fig. \ref{fig:Fig2EffectPoseError}b). 
By ignoring perspective distortions, the same image plane error caused by $\Delta \beta$ can also be caused by a corresponding (adjusted) $\Delta t_{x}$ (Fig.~\ref{fig:Fig2EffectPoseError}c). Thus, both errors in $\Delta \beta$ and $\Delta t_{x}$ can be compensated by a single correction either in $\Delta \beta$ or $\Delta t_{x}$. %
This also applies for $\Delta \alpha$ and $\Delta t_{y}$. %
The correspondence between correlating $\Delta t_{x}$, $\Delta t_{y}$ and $\Delta \alpha$, $\Delta \beta$ is unique and constant for all points. This does not apply for $\Delta \gamma$, which would require an individual $\Delta t_{x}$,$\Delta t_{y}$ compensation for each point (Fig. \ref{fig:Fig2EffectPoseError}c). \vspace{-0.8em}
This analysis suggests, that three instead of six extrinsic parameters are sufficient for pose error reduction in our case. Since $\Delta t_{z}$ has little effect and $\Delta \alpha$,$\Delta \beta$ can be compensated, only $\Delta t_{x}$, $\Delta t_{y}$, $\Delta \gamma$ need to be estimated. Less parameters for pose error estimation leads to significantly faster and more efficient optimizations.    \vspace{-0.8em}   
\section{Efficient Image Integration}
In \cite{Kurmi2020GeoScience} we show that the GLV \cite{Firestone1991} of an AOS thermal integral image $X$ is an occlusion invariant measure for determining the optimal synthetic focal plane pose, which minimizes occlusion and defocus. We prove (see Appendix) that the variance of an integral image is proportional to
\begin{equation}\label{Eq:VarianceRelationMain}
  \Var[\X] = \frac{ \Var[\X_{i}]}{\N} + \OneD^2\big(1-\frac{1}{\N}\big)\sigmaSign,
\end{equation}
where the variance of a single image is 
\begin{equation}\label{Eq:IndividualImageVarMain}
  \Var[\X_i] = \p\OneD((\meanOccl - \meanSign)^{2}) + \p\sigmaOccl + \OneD\sigmaSign, 
\end{equation}
and $\p$ is the probability of occlusion, while $\meanOccl$, $\sigmaOccl$  and $\meanSign$, $\sigmaSign$ are statistical properties of occlusion and the signal respectively. 
Note that the variance of a single image $\Var[\X_i]$ is only dependening on its content (signal and occlusion). %
Individual pose parameters ($\alpha, \beta, \gamma, t_x, t_y,t_z$) do not influence $\Var[\X_i]$ but indirectly influence the focus/defocus of the integral image $\X$. %
Thus, integrating $N$ individual images that are miss-registered due to remaining errors in pose estimation will result in defocus and thus change the variance of the integral image $\Var(X)$---even if the optimal focal plane settings are found.
\begin{figure*}[t!]
    \centering
    \includegraphics[width=0.98\linewidth]{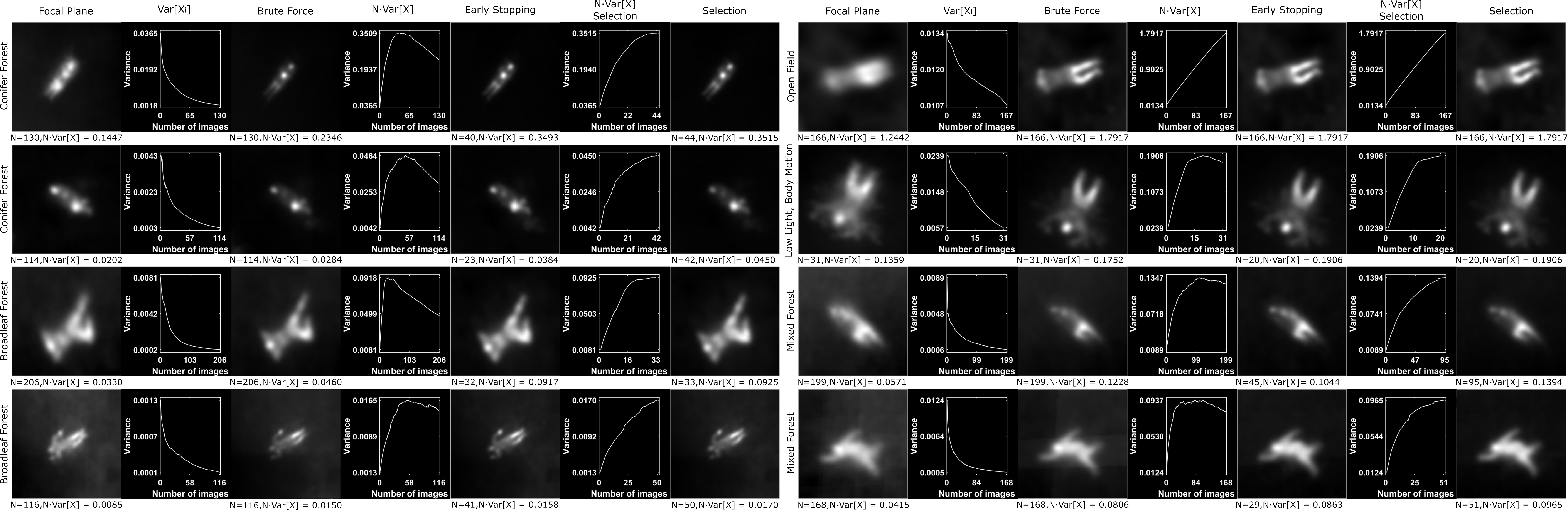}
    \vspace{-0.4em}\caption{Integration strategies for a set of representative cases (different scene types shown in Fig. \ref{fig:Fig1Completeresult}). From left to right: Results of {\it focal plane optimization} only (without additional pose error reduction), $\Var[\X_i]$-plot for all individual images in variance-decreasing order, results of {\it brute force} integration of all images in variance decreasing order, $N \cdot \Var[\X]$-plot for integral images computed from $N$ individual images in variance-decreasing order, results of {\it early stopping} the integration (in variance-decreasing order) at the maximum of $N \cdot \Var[\X]$, $N \cdot \Var[\X]$-plot for images that lead to an increase of $N \cdot \Var[\X]$ after integration in variance-decreasing order ({\it selection}), results of the {\it selection} strategy.} 
    \label{fig:Fig3IntegrationStrategies}
    \vspace{-1.7em}
\end{figure*}
We propose to optimize the initial pose estimation results of each image $X_i$ by maximizing $\Var[\X]$ while searching for better pose parameters $(\alpha,\beta,\gamma,t_x,t_y,t_z)_i$, where $\X$ is the integral image of all $X_i$ ($i=1 \dots N$). 
First, we determine the optimal synthetic focal plane, as explained in \cite{Kurmi2020GeoScience}. As explained earlier, we use general-purpose structure-from-motion and multi-view stereo pipeline \cite{schonberger2016sfm,schoenberger2016mvs} applied to the high resolution RGB images for initial pose estimation. 
Second, we sort all thermal images $X_i$ in decreasing $\Var[\X_i]$ order, since higher variance indicates less occlusion. 
Finally, we sequentially register and integrate all images $\X_i$ in decreasing variance order to an integral image $\X$ by maximizing $\Var[\X]$ while searching for optimal pose parameters $(\alpha,\beta,\gamma,t_x,t_y,t_z)_i$ which register $\X_i$ to the previous integral image (i.e., the one that integrates $\X_1 \dots \X_{(i-1)}$) -- starting initially with $\X=\X_1$. This way, we have to optimize only for six parameters per integration step, rather than for $6N$ parameters at once. %
We chose Nelder-Mead \cite{Nelder1965} (a commonly used direct search approach with complexity of $\mathcal{O}(n)$ per iteration where $n$ is the number of parameters) for parameter optimization.
Note, that when integrating $N$ images, the contrast of the resulting integral image drops proportionally to $N$ (due to the present occlusion on each individual image)\cite{Kurmi2019Sensor}. Thus, $N \cdot \Var[\X]$ normalizes the resulting variance to become invariant to $N$.

As shown in Fig. \ref{fig:Fig3IntegrationStrategies}, this clearly improves the focus of the integral images compared to a focal plane optimization only. %
However, such a brute force registration of all $N$ images by optimizing all six parameters is still time consuming and scales linearly with $N$, even if we search only $N$ times within a 6-dimensional parameter space rather than once within a $6N$-dimensional parameter space. Furthermore, focus can worsen if too many images are integrated in case of remaining optimization errors.

To overcome these problems, we investigated two alternatives to the above \textit{brute force} strategy: First, an \textit{early stopping} strategy stops the integration process as soon as $N \cdot \Var[\X]$ has reached its maximum and starts to decrease. Second, an \textit{selection} strategy registers all images but integrates only those that lead to an increase in $N \cdot \Var[\X]$. Results of these three integration strategies are illustrated in Fig. \ref{fig:Fig3IntegrationStrategies}.  

Since early stopping is fastest and does not show a significant degradation in focus for a set of representative test cases, we chose this as a final integration strategy.  \vspace{-1.5em}   
\section{Search Space Reduction}

As discussed in section \ref{section:EffectofPoseErrors}, we could further reduce the number of pose parameters to be optimized for each image from six ($t_x,t_y,t_z,\alpha,\beta,\gamma$) to three ($t_x,t_y,\gamma$). Here, we evaluate this option for our representative test cases while applying early stopping for integration. 

\begin{figure*}[!t]
    \centering
    \includegraphics[width=0.98\linewidth]{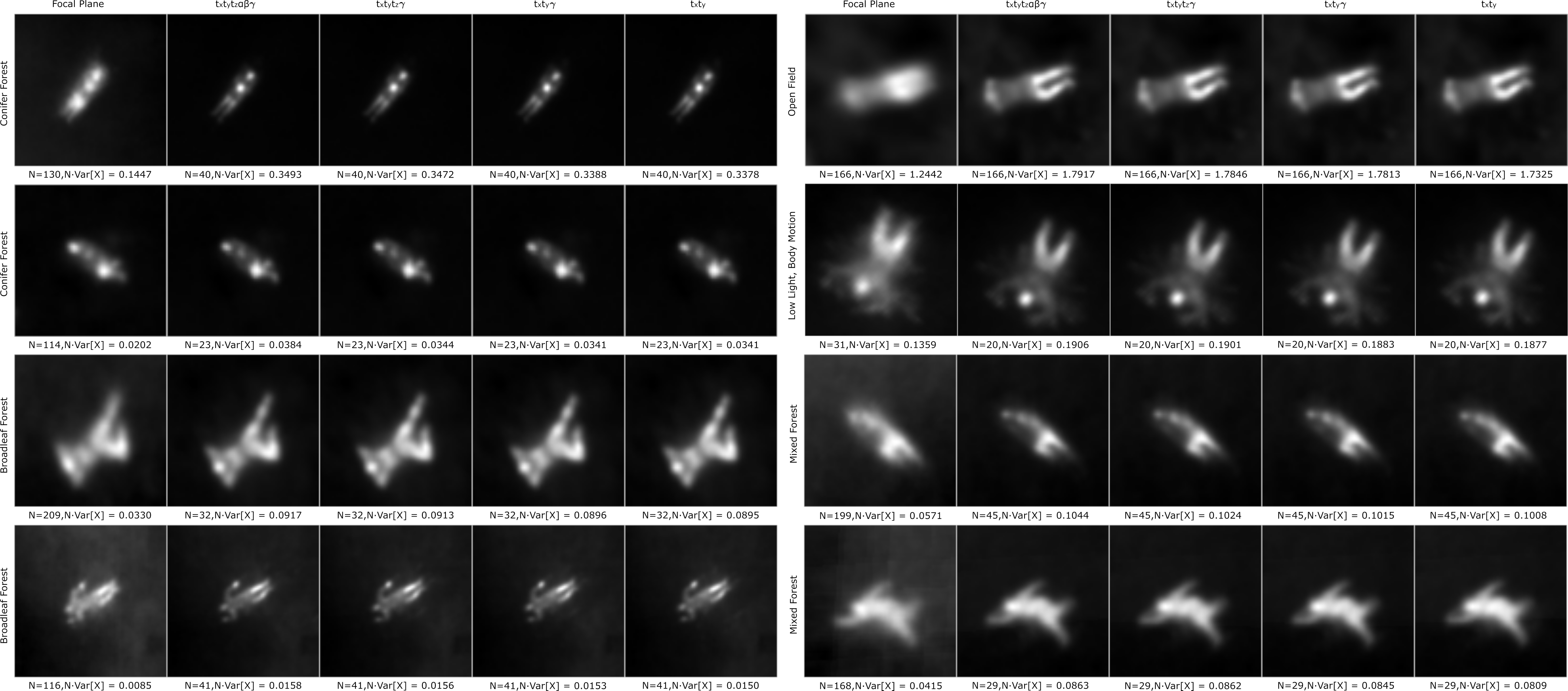}
   \caption{Search space reduction by decreasing the number of pose parameters for a set of representative cases (same as in Fig. \ref{fig:Fig3IntegrationStrategies}). From left to right: Results of {\it focal plane optimization} only (without additional pose error reduction), {\it early stopping} integration applied to all six parameters ($t_x,t_y,t_z,\alpha,\beta,\gamma$), applied to four parameters ($t_x,t_y,t_z,\gamma$), applied to three parameters ($t_x,t_y,\gamma$), and applied to two parameters ($t_x,t_y$).}
    \label{fig:Fig4SearchSpace}
    \vspace{-1.6em}
\end{figure*}

As illustrated in Fig. \ref{fig:Fig4SearchSpace}, there is no considerable degradation in focus even if the number of pose parameters is reduced to two ($t_x,t_y$). In section \ref{section:EffectofPoseErrors} we explain that errors in $\alpha,\beta$ can be compensated by corrections in $t_x,t_y$, and that $t_z$ has no significant impact in case of $t_{z}\gg Z$. 
The reason why corrections of $\gamma$ seems also to have little impact might be that the remaining error resulting from the initial pose estimation was little for our test cases. However, since this cannot be assumed in general, we chose the set of three pose ($t_x,t_y,\gamma$) as the best trade-off between quality and performance. 
\vspace{-0.5em}
\vspace{-0.5em}
\section{Results}
Figure \ref{fig:Fig1Completeresult} presents final results of our pose error reduction approach being applied to a total of 52 test cases captured under different conditions (conifer forest, broadleaf forest, mixed forest, open field, low-light and sunlight conditions, moving and still subjects, different densities). It achieves significant qualitative improvements under open field conditions (without occlusion), sparse and moderate dense occlusion conditions, and even under low-light conditions with body motion. Under conditions with severe occlusion, however, only minor improvements are made.

Our optimization framework (including search space reduction and early stopping) has, on an average, $89.92\%$ less parameters to optimize when compared to brute force search (only $3N$ instead of $6N$ parameters, where on average $N$ is lower due to early stopping).
It significantly improves the focus of the integral images by $204.77\%$ (according to the measured $N \cdot \Var[\X]$ gain) when compared to only optimizing the focal plane as in \cite{Kurmi2020GeoScience}.

When executed on a Nvidia GeForce GTX 1060, our GPU implementation requires on average 355s for the optimization of AOS integrals computed form 50 single images (before early stopping). \vspace{-0.8em}
\section{Conclusion and Future Work}
In this article we utilize GLV as an optimization metric for pose error reduction in thermal AOS integral images. We study effects of errors in different pose parameters and demonstrate that the number of pose parameters to be optimized can be reduced from six to three without any significant degradation of quality. We also investigate different image integration strategies to further reduce the search space, and showed that the contribution of a single image in the total integral is proportional to its GLV. Thus, single images are best integrated in decreasing GLV order -- which supports early stopping.

Our results demonstrate that we achieved qualitative and quantitative improvements for representative test cases. The reported computation time of our implementation (on average 355s) can be reduced by an efficient hierarchical multi-scale processing approach (with a first test implementation we already achieved an average of 197s for each test case). Furthermore, advanced high-end GPUs (e.g., Nvidia Tesla V100 or Quadro RTX 8000) will lead to $\times$3-4 speed-ups when compared to the applied Nvidia GeForce GTX 1060. Overall, we estimate that processing times of less than 50s will be realistic and applicable as post-processing image-enhancement step. In comparison, conventional PnP solutions, such as in \cite{schoenberger2016mvs,schonberger2016sfm}, required on average 24 minutes for pose estimations with the full set of around 300 images.

In the future, we want to investigate if our pose-error-reduction approach (based on thermal images) can replace the conventional pose-estimation process (based on RGB images).
The latter is not applicable at night as it relies on sufficient visible light. If the drone's GPS recordings are used as initial poses and our approach reduces the remaining pose errors sufficiently, then AOS night operations are possible. 

We are currently also working on a people classification deep neural network that operates on unoptimized AOS integral images. This will allow the fully automatic selection of people regions to be optimized (which are manually selected at the moment).  
\vspace{-1.3em}
\section*{Appendix}
Here, we present the derivation of the relation between variance ($\Var[X]$) of the integral image $\X$ and variance ($\Var[X_i]$) of the individual image recordings $\X_i$. We rely on the statistical model of \cite{Kurmi2020GeoScience}, where the integral image $\X$ is composed of $N$ single image recordings $\X_i$ and each single image pixel in $\X_i$ is either occlusion free ($\img$) or occluded ($\noise$) determined by $Z$:
\appendixMath{\begin{equation}\label{eq:IndividualImageComp}
  \X_{i} = Z_{i}\noise_{i}  + (1 - Z_{i})\img .
\end{equation}}
Similar to \cite{Kurmi2020GeoScience}, all variables are independent and identically distributed with $Z_{i}$ following a Bernoulli distribution with success parameter $\p$ (i.e.,
$\Ex{Z_i}=\Ex{Z_i^2}=\p$; furthermore, note that $\E[Z_{i}(1-Z_{i} )] = 0$ is true). %
The random variable $\img$ follows a distribution with mean $\Ex{\img}=\meanSign$ and $\Ex{\img^2}=(\meanSign^2+\sigmaSign)$ and analogously $\noise_{i}$ follows a distribution with mean $\Ex{\noise_i}=\meanOccl$ and $\Ex{\noise_i^2}=(\meanOccl^2+\sigmaOccl)$. %
For the mean and the variance of $\X_i$, we determine the first
and second moments of $\X_i$:
\appendixMath{\begin{equation}
\label{eq:IndividualImageMean} 
\Ex{\X_i} = \p\meanOccl + \OneD\meanSign %
\end{equation}}
and 
\appendixMath{\begin{equation}
\label{eq:IndividualImagePower} 
\Ex{{\X_i}^2} =  \p(\meanOccl^2+\sigmaOccl) + \OneD(\meanSign^2+\sigmaSign). %
\end{equation}}
Variance of single image recordings $\X_i$ can be obtained as:
\appendixMath{\begin{equation}\label{Eq:IndividualImageVariance}\begin{split}
  \Var[\X_{i}] &= \Ex{\X_{i}^2} - (\Ex{\X_{i}})^{2}\\
  &= \p(\meanOccl^2+\sigmaOccl) + \OneD(\meanSign^2+\sigmaSign) - (\p^{2}\meanOccl^{2} + \OneD^{2}\meanSign^{2} + 2\p\OneD\meanOccl\meanSign)\\
&= \p\OneD((\meanOccl - \meanSign)^{2}) + \p\sigmaOccl + \OneD\sigmaSign.
\end{split}\end{equation}}
Similarly, for $\X$, we determine the first
and second moments where the first moment of $\X$ is given by:
\appendixMath{\begin{equation}\begin{split}
  \Ex{\X} &= \E\left[\frac{1}{\N} \sum_{i=1}^N Z_{i} \noise_i + \OneZi\R\right]\\
  &= \p \meanOccl + \OneD\meanSign
\end{split}\end{equation}}
and the second moment of $\X$ is as derived in \cite{Kurmi2020GeoScience}:
\appendixMath{\begin{equation}\begin{split} \label{eq:EX2}
\Ex{\X^2}  &= %
\frac{1}{\N^2}\bigg(\N \big( \p(\sigmaOccl + \meanOccl^2) + \OneD (\sigmaSign +\meanSign^2)\big) %
\\ &  + {N(\N\!-\!1)} \big( \p^2\meanOccl^2 + 2\p \OneD \meanSign \meanOccl +\OneD^2(\sigmaSign + \meanSign^2) \big) \bigg). \\
\end{split}\end{equation}}
Consecutively, we calculate variance of the integral image as:
\appendixMath{\begin{equation}\label{Eq:IntegratedlImageVariance}\begin{split}
  \Var[\X] &= \Ex{\X^2} - (\Ex{\X})^2\\
  &= \frac{1}{\N} \big( \p(\sigmaOccl + \meanOccl^2) + \OneD (\sigmaSign +\meanSign^2)\big) %
\\ & \quad + \big( \p^2\meanOccl^2 + 2\p \OneD \meanSign \meanOccl +\OneD^2(\sigmaSign + \meanSign^2) \big) 
\\&\quad - \frac{1}{\N}\big( \p^2\meanOccl^2 + 2\p \OneD \meanSign \meanOccl +\OneD^2(\sigmaSign + \meanSign^2) \big) 
  \\&\quad - \big(\p^{2}\meanOccl^{2} + \OneD^{2}\meanSign^{2} + 2\p\OneD\meanOccl\meanSign\big)\\
  &= \frac{1}{\N} \big( \p(\sigmaOccl + \meanOccl^2) + \OneD (\sigmaSign +\meanSign^2)\big) %
\\ & \quad + \OneD^2\sigmaSign
\\&\quad - \frac{1}{\N}\big( \p^2\meanOccl^2 + 2\p \OneD \meanSign \meanOccl +\OneD^2(\sigmaSign + \meanSign^2) \big)\\
   &= \frac{1}{\N} \big(\p\OneD((\meanOccl - \meanSign)^{2}) + \p\sigmaOccl + \OneD\sigmaSign \big)
  \\ & \quad + \OneD^2\big(1-\frac{1}{\N}\big)\sigmaSign.
\end{split}\end{equation}}
Substituting (\ref{Eq:IndividualImageVariance}) in (\ref{Eq:IntegratedlImageVariance}) yields   (\ref{Eq:VarianceRelationMain}).

\ifCLASSOPTIONcaptionsoff
  \newpage
\fi

\printbibliography
\end{document}